# Extraction and Analysis of Clinically Important Follow-up Recommendations in a Large Radiology Dataset

Wilson Lau[1], Thomas H. Payne, MD[2,3], Ozlem Uzuner, PhD[4], Meliha Yetisgen, PhD[1]
[1]Department of Biomedical and Health Informatics, [2]School of Medicine, [3]Information Technology Services, University of Washington, Seattle, WA
[4]Department of Information Sciences and Technology, George Mason University, Fairfax, VA

**Abstract**

*Communication of follow-up recommendations when abnormalities are identified on imaging studies is prone to error. In this paper, we present a natural language processing approach based on deep learning to automatically identify clinically important recommendations in radiology reports. Our approach first identifies the recommendation sentences and then extracts reason, test, and time frame of the identified recommendations. To train our extraction models, we created a corpus of 567 radiology reports annotated for recommendation information. Our extraction models achieved 0.92 f-score for recommendation sentence, 0.65 f-score for reason, 0.73 f-score for test, and 0.84 f-score for time frame. We applied the extraction models to a set of over 3.3 million radiology reports and analyzed the adherence of follow-up recommendations.*

**Introduction**

With the dramatic rise in utilization of medical imaging in the past two decades, health providers are challenged by the optimal use of clinical information while not being overwhelmed by it. A radiology report is the principal means by which radiologists communicate the findings of an imaging test to the referring physician and sometimes the patient. Based on potentially important observations in the images, radiologists may recommend further imaging tests or a clinical follow-up in the narrative radiology report. These recommendations are made for several potential reasons: (1) radiologists may recommend further investigation to clarify the diagnosis or exclude potentially serious, but clinically expected disease; (2) radiologists may unexpectedly encounter signs of a potentially serious disease on the imaging study that they believe require further investigation; (3) radiologists may recommend surveillance of a disease to ensure an indolent course; or (4) they may provide advice to the referring physician about the most effective future tests specific to the patient's disease or risk factors. The reliance on human communication, documentation, and manual follow-up is a critical barrier to ensuring that appropriate imaging or clinical follow-up occurs. There are many potential points of failure when communicating and following up on important radiologic findings and recommendations: (1) <u>Critical findings and follow-up recommendations not explicitly highlighted by radiologists</u>: Although radiologists describe important incidental observations in reports, they may or may not phone an ordering physician. If these recommendations "fall through the cracks", patients may present months later with advanced disease (e.g., metastatic cancer). (2) <u>Patient mobility</u>: When patients move between services in healthcare facilities, there is increased risk during "hand-offs" of problems with follow-up of test result and continuity of care[1]. (3) <u>Heavy workload of providers</u>: Physicians and other providers have to deal with a deluge of test results. A survey of 262 physicians at 15 internal medicine practices found that physicians spend on average 74 minutes per clinical day managing test results, and 83% of physicians reported at least one delay in reviewing test results in the previous two months[2]. However, it is vital that these results, particularly if they are unexpected, are not lost to follow-up. In patients who have an unexpected finding on a chest radiograph, approximately 16% will eventually be diagnosed with a malignant neoplasm[3].

These examples indicate an opportunity to develop a systematic approach to augmenting existing channels of clinical information for preventing delays in diagnosis. The goals of our research are to: (1) build a gold standard corpus of radiology reports annotated with recommendation information, (2) build information extraction approaches based on deep learning to automatically identify recommendation information, and (3) apply the trained extractors to a large dataset of 3.3 million radiology reports created at University of Washington and Harborview Medical Center between 2008 and 2018 to analyze follow-up adherence rates.

In this research, we define a *follow-up recommendation* as a statement made by the radiologist in a given radiology report to advise the referring clinician to further evaluate an imaging finding by either other tests or further imaging. Figure 1 presents a radiology report with such a follow-up recommendation. In our annotation, we first labeled sentences with recommendation. For each identified recommendation, we also annotated the spans that describe (1) the reason for follow-up, (2) recommended test, and (3) time frame. In figure 1, the recommendation sentence is

"*Given family history, would recommend repeat ultrasound in 4-5 weeks to evaluate fetal growth and complete anatomic survey*"; reason is "*to evaluate fetal growth and complete anatomic survey*"; recommended test is "*ultrasound*", and time frame is "*4-5 weeks*".

> IMPRESSION
>
> Singleton pregnancy. Size consistent with dates. Anatomic survey limited by maternal body habitus and fetal position. Inadequate views of fetal heart and spine. *Given family history, would recommend repeat ultrasound in 4-5 weeks to evaluate fetal growth and complete anatomic survey*. If unable to visualize fetal heart at that time, consider fetal echo.

**Figure 1.** Example radiology report with recommendation information annotations.

## Related Work

Automated information extraction using natural language processing (NLP) techniques has made patient information in clinical notes accessible for scientific research. Informatics for Integrating the Biology and the Bedside (i2b2) has been organizing NLP challenges on different types of clinical information extraction since 2006. These challenges included private health information de-identification[4], medical concept extraction[5], temporal information extraction[6], as well as medication information extraction[7]. Participants employed different NLP approaches including rule-based, machine learning and ensemble methods to address these tasks. Machine learning methods were usually based on statistical classification algorithms such as Support Vector Machines (SVMs), Maximum Entropy (MaxEnt) and Conditional Random Fields (CRFs)[8,9]. In recent years, neural networks have gained tremendous popularity, especially after several breakthroughs were accomplished by Hinton and Mikolov, T. et al.[10,11] and several deep learning libraries became publicly available. Clinical NLP researchers have also taken this opportunity and employed neural network modeling to deliver state-of-the-art performance. For instance, the best de-identification system in 2009 achieving 98% F-score was based on statistical learning with regular expression[12]. In 2016, Dernoncourt et al. were able to achieve similar performance using bidirectional LSTM and character embedding[13].

Prior research efforts on radiology follow-up recommendation detection were primarily based on rule-based and machine learning approaches. Dutta et al. employed lexicons matching heuristics to detect recommendations for incidental findings[14]. They compiled a set of lexicons which consisted of various inflectional morphemes of the same stem words. They went through three iterations of development and validation to fine tune their pattern matching algorithm. Chapman et al. adopted an algorithm, pyConTextNLP, to identify critical findings from radiology reports that were relevant to abdomen, chest, neuro and spine exams[15]. The algorithm used classification rules that were based on specific sentence structures in the reports. It also relied on a knowledge base that captured common biomedical terms in the target radiology imaging reports. Johnson et al. evaluated the ConText algorithm with a chest X-ray report and found that the algorithm misidentified two cases of negation and temporality in three sentences[16]. They proposed a heuristic approach to identify incidental findings based on regular expressions with patterns of lexicons. They argued that their approach could outperform processes that solely relied on radiologist annotations. However, their evaluation was based on a small and highly imbalanced dataset of 580 records (8.6% positive to negative ratio) and was only limited to X-ray, CT and ultrasound. Another lexicon based commercial system, LEXIMER (Lexicon Mediated Entropy Reduction), was used by Dang et al. to identify recommendations across different modalities. These authors analyzed the results using OLAP (Online analytical processing) technologies, a common approach in business intelligence and data warehousing[17,18]. LEXIMER parsed the reports into phrases and then weighted the phrases using hierarchical decision trees against a dictionary of lexicons[19]. Similarly, Mabotuwana et al extracted follow-up recommendations and associated anatomy from radiology reports with a keyword-based heuristic approach to identify recommendations in finding and impression sections of over 400 thousand radiology reports[20]. The same group processed close to 3 million radiology notes to determine adherence rates to follow-up recommendations[21].

Domain adaptability is a major problem for rule-based and lexicon-based approaches as these methods require expert intervention to upkeep the logic of the rules and the dictionaries, which are often tailored to a specific problem and/or domain. Statistical NLP methods on the other hand do not require domain expert maintenance since the model automatically learns from annotated examples. Yetisgen-Yildiz et al. developed a Maximum Entropy classifier for recommendation detection that achieved a best F-score of 87% based on a very rich set of features including ngrams, UMLS concepts, syntactic, temporal as well as structural features[22]. Martinez et al. employed a similar machine learning approach to identify sentences associated with invasive fungal diseases (IFD) in CT scan reports[23]. They developed two different binary SVM classifiers to classify three different labels, i.e., IFD positive vs.

IFD neutral and IFD negative vs. IFD neutral. Their feature set included part-of-speech tags, UMLS concepts, as well as word sense disambiguation and negation indicators. Their IFD positive classifier achieved a best F-score of 70.5% while the IFD negative classifier achieved a best F-score of 77.2%.

Deep learning is not restricted by the lengthy process of handcrafted feature engineering usually required by traditional statistical NLP approaches for better performance. Instead, intricate distributed features are learned by adjusting model weights through backpropagation. Traditional methods suffer from word sense ambiguity and out-of-vocabulary tokens in clinical text which often contains misspellings, acronyms and foreign words. A common solution would be using dictionary of lexicons and gazetteers[24]. Deep learning can overcome these issues by the notion of transfer learning where the model is first trained on a larger dataset in a similar context and then fine-tuned on a smaller target dataset with limited number of annotated labels[25]. Another approach is to use character embeddings where the model is able to learn the morphological features of words, such as prefixes, suffixes, and any sub-token patterns to account for out-of-vocabulary words.

In this paper, we present a deep learning NLP system for extracting recommendation information from radiology imaging reports. We first develop a binary classifier based on Hierarchical Attention Networks[26] to identify follow-up recommendation sentences and then apply a state-of-the-art deep neural named entity extraction system NeuroNER[27] to extract three entities: reason, test, time frame. These attributes help us understand why a follow-up recommendation is made by a radiologist to advise referring clinician for further evaluation. To our knowledge, this will be the first study that applied deep learning to the problem of recommendation extraction in a dataset of 3.3 million radiology reports.

## Methods

**Datasets:**

**Pilot Corpus:** In previous work, we created a corpus composed of 800 de-identified radiology reports extracted from the radiology information system of our institution. The reports represented a mixture of imaging modalities, including radiography computer tomography (CT), ultrasound, and magnetic resonance imaging (MRI). The distribution of the reports across imaging modalities is listed in Table 1.

| Imaging modality | Number of reports |
|---|---|
| Computer tomography | 486 |
| Radiograph | 259 |
| Magnetic resonance imaging | 45 |
| Ultrasound | 10 |
| Total | 800 |

Table 1. Distribution of reports in pilot corpus.

Annotation Guidelines: We annotated this dataset prior to defining different categories of follow-up recommendations. In this annotation task, we asked the annotators simply to highlight the boundaries of sentences that include any follow-up recommendations.

Annotation Process: Two annotators, one radiologist and one internal medicine specialist, went through each of the 800 reports independently and marked the sentences that contained follow-up recommendations. Out of 18,748 sentences in 800 reports, the radiologist annotated 118 sentences and the clinician annotated 114 sentences as recommendation. They agreed on 113 of the sentences annotated as recommendation. The inter-rater agreement measured in terms of F-score was 0.974.

**Multi-institutional Radiology Corpus:** We extended our pilot dataset of 800 reports with a much larger set of 3,301,748 radiology reports from two different institutions including the University of Washington Medical Center (1,903,772 reports) and Harborview Medical Center (1,397,976 reports) from year 2008 to 2018. University of Washington Human Subjects Division Institutional Review Board approved retrospective review of this dataset. Table 2 shows the distribution of radiology reports by modality in this larger dataset.

| Imaging Modality | Number of reports |
|---|---|
| Angiography | 53,658 |
| Computed Tomography | 706,908 |
| Fluoroscopy | 1,072 |
| Magnetic Resonance Imaging | 243,833 |
| Mammogram | 157,374 |
| Nuclear Medicine | 58,350 |
| Portable Radiography | 310,311 |
| Positron emission tomography | 1,799 |
| Ultrasound | 351,761 |
| X-Ray | 1,416,682 |
| Total | 3,301,748 |

**Table 2.** Distribution of reports in multi-institutional radiology corpus

Annotation Process: We designed the annotation task to operate on two levels: sentence level and entity level. At the sentence level, the annotators mark the boundaries of recommendation sentences. At the entity level, the annotators mark three attributes of recommendation information presented in the marked sentences: (1) Test: the imaging test or clinical exam that is recommended for follow-up, e.g., *screening breast MRI or CT*, (2) Time frame: the recommended time frame for the recommended follow-up test or exam, e.g., *1-3 weeks, 12 months*, and (3) Reason: the reason for the critical follow-up recommendation, e.g., *to assess the actual risk of Down's Syndrome*.

Because manual annotation is a time-consuming and labor-intensive process, we could annotate only a small portion of our large radiology corpus. The percentage of reports that include recommendation sentences is quite low—about 15% at our institution. To increase the number of reports with recommendations in the annotated set, rather than randomly sampling, we built a high recall (0.90), low precision (0.35) classifier trained on the pilot dataset. The details of this baseline classifier can be found in our prior publication[28]. We ran our baseline classifier on un-annotated reports and only sampled from the ones identified as positive by our classifier for manual annotation. Because the baseline classification was high recall and low precision, the false positive reports could subsequently be corrected by our annotators. The filtering of reports using a classifier reduced the number of reports that our human annotators needed to review, thereby expediting the annotation process.

At the sentence level, one radiologist and one neurologist reviewed the classifier-selected reports with system generated follow-up recommendation sentences. The annotators corrected the system generated sentences and/or highlighted new sentences if needed.

At the entity level, one neurologist and one medical school student annotated the entities (reason for recommendation, recommended test, and time frame) in reports annotated in a previous stage at the sentence level with follow-up recommendations.

Inter-annotator Agreement Levels: At the sentence level, we measured the inter-annotator agreement on a set of 50 reports featuring at least one system-generated recommendation identified by our high recall classifier from a randomly selected collection of one thousand reports. Our annotation process required annotators to go over all sentences that were initially identified by the system as a recommendation. They could label the sentence as *Incorrect* if they believed the system had wrongly identified a recommendation sentence (false positive) or if they believed the system had missed the sentence (false negative). The inter-rater agreement levels were kappa 0.43 and 0.59 F1 score for the first iteration. To resolve the disagreements, we scheduled multiple meetings. One of our observations during those meetings was that none of the new recommendation sentences introduced by either annotator were identified by the other. In our review, both annotators agreed that the majority of the new recommendations introduced by the other were correct. We adjusted our annotation guidelines to add rules to help decide if and when a new sentence should be identified as a recommendation.

At the entity level, agreement levels were 0.78 F1 for reason, 0.88 F1 for test, and 0.84 F1 for time frame. Our final annotated corpus contained 597 positive examples of recommendation sentences and 11787 negative examples of recommendation sentences from 567 radiology reports. At the entity level there were 735 test, 173 time frame and 545 reason entities in the final corpus.

**Approach:**

<u>Recommendation extraction</u>: To identify sentences that include recommendation information, we first chunk reports into sentences with NLTK sentence tokenizer. Table 3 shows the distribution of sentences by image modality on the multi-institutional radiology dataset. As can be seen in the table, the length of reports varies across modalities.

| Imaging Modality | Number of sentences | Average number of sentences per report |
|---|---|---|
| Angiography | 1,504,939 | 28.05 |
| Computed Tomography | 18,109,590 | 25.62 |
| Fluoroscopy | 13,452 | 12.55 |
| Magnetic Resonance Imaging | 5,688,512 | 23.33 |
| Mammogram | 2,016,911 | 12.82 |
| Nuclear Medicine | 1,144,518 | 19.62 |
| Portable Radiography | 2,055,534 | 6.62 |
| Positron emission tomography | 41,423 | 23.03 |
| Ultrasound | 6,841,966 | 19.45 |
| X-Ray | 10,008,031 | 7.06 |

**Table 3.** Distribution of sentences by image modality in the multi-institutional radiology corpus

We defined our recommendation sentence extraction task as a classification problem at the sentence level. We implemented our sentence classifier based on Hierarchical Attention Networks (HANs)[26]. HAN is a neural model that employs a stacked recurrent neural network architecture. In particular, the weights of the hidden layers for each word are aggregated by an attention mechanism to form a sentence vector. The importance of each word in association with the outcome label is represented by the attention weight vector that can be learned by a word encoder which is implemented as a layer of bidirectional Gated Recurrent Unit (GRU). The attention weight vector α is computed through a softmax function of the input context vector and a single hidden layer. Intuitively, the attention vector represents how important the word is in determining the outcome label. The sentence vector which is made up of these word attentions are then passed to another similar attention mechanism where the importance of sentences can also be learned by another layer of bidirectional GRU (sentence encoder). The bidirectional nature of the encoders allows the contextual information in the input to be read in both directions and summarized. The hierarchical architecture allows the model to learn the context of a document by summarizing the context of its sentences, each of which in turn was summarized by its own words. The ability to selectively learn from local segments of texts to predict the outcome labels is a unique characteristic of the attention mechanism in deep learning. This network model has been proven to be more effective than conventional statistical machine learning approaches in extracting clinical information[29]. Since radiology recommendations follow similar hierarchical structure which consist of multiple sentences made up of multiple words, the HAN model is suitable for our recommendation classification task.

Hyperparameter optimization: We pretrained our word embeddings using Word2Vec on the entire 10 years of radiology dataset. We used grid search to find the best set of hyperparameters. Based on our preliminary experiments, we identified the range for each hyperparameter in the search space, which was also limited by available system memory: Word2Vec embedding dimension (100-300); number of bidirectional GRU unit on word encoder (100 - 500); number of bidirectional GRU unit on sentence encoder (100 - 500); drop out (0.3 - 0.5). We have also experimented with both Adam and SGD optimizers. Table 5 shows our best hyperparameter configuration.

| Parameter | Value |
|---|---|
| word2vec embedding dimension | 300 |
| number of bidirectional GRU unit on word encoder | 300 |
| number of bidirectional GRU unit on sentence encoder | 300 |
| drop out | 0.4 |
| optimizer | Adam |

**Table 4**. HAN hyperparameter configuration

We used 0.9/0.1(train/validation) split on the training dataset. We applied early stopping technique[30] to avoid overfitting with patience level set to 15 epochs. On each epoch, we evaluated the model based on the predicted F1 score on the validation set. The training would stop when no improvement was shown in the last 15 epochs.

Named Entity Recognition: We used Dernoncourt et al.'s NeuoNER[27] to process our annotated files in BRAT standoff format. The core of NeuroNER consists of two stacked layers of recurrent neural networks. The first layer is the *Character-enhanced token-embedding layer* in which the embedding of each word token is learned by a bidirectional LSTM from its character embedding. The resulting token embedding is then concatenated with our pretrained Word2Vec word embeddings to form an enhanced token embedding. These token embeddings are then processed by another layer of bidirectional LSTM, the *Label prediction layer*, to compute the probability vector of each word token being one of the entities. Finally, the sequence of probability vectors is sent to a feed-forward layer, the *Label sequence optimization layer*, to determine the predicted entity for each token by taking argmax of the probability vector, i.e., the entity label with the highest probability for each token. The character embedding captures the morphological features of word tokens and performs particularly well in handling morphemes, acronyms, misspellings and out-of-vocabulary tokens. It provides another level of word presentation that is not captured by sampling word co-occurrence as in Word2Vec and GloVe. This network architecture achieved state-of-the-art performance in identifying PHI information in i2b2 dataset and MIMIC dataset[13].

We used BIOES annotation (Begin, Inside, Outside, End, Single) to tag each token in the sequence and performed 5-fold cross validation on the training corpus. We pretrained our own word embeddings with Word2Vec on the multi-institutional radiology corpus of 3.3 million radiology reports.

## Results

Recommendation extraction: Our training corpus contains 597 positive sentences and 11787 negative sentences from 567 radiology reports. Our recommendation extraction model based on deep learning achieved 0.88 precision, 0.96 recall, and 0.92 f-score with 5-fold cross validation (true positive: 574, true negative: 11711, false positive: 75, false negative: 22). In previous work, for the same problem, we achieved 0.66 precision, 0.88 recall, 0.76 f-score with Max-Ent classifier with extensive feature engineering[28].

Named-entity recognition: Table 5 shows the token-based 5-fold cross validation results on the three entities.

| Entity | Precision | Recall | F1 |
|---|---|---|---|
| Reason | 68.53 | 62.05 | 65.10 |
| Test | 74.20 | 71.48 | 72.71 |
| Time frame | 83.38 | 85.05 | 84.16 |

**Table 5.** Token level entity extraction 5-fold cross-validation results

Analysis of multi-institutional dataset: The recommendation extraction model predicted 871,680 recommendations in the total of 47,424,876 sentences. Table 6 shows the distribution of predicted recommendations and table 7 presents examples of extracted recommendation sentences by modality. 658,303 reports (19.9%) in the entire dataset included recommendations. As can be observed from Table 6, 98.7% of mammograms included a follow-up exam. For other modalities, percentages of reports with recommendations varied between 6.04% (portable radiography) and 30.74% (ultrasound). To evaluate the performance of our recommendation extraction model, we randomly selected 40 recommendations for top 5 modalities with highest recommendation percentages: mammograms (98.70%), ultrasound (30.74%), positron emission tomography (28.79%), computed tomography (26.38%), and angiography (24%) and manually validated their correctness. We identified 178 out of 200 those recommendations as true positives which resulted a precision value (0.89) on the target dataset similar to our 5-fold cross validation result (0.88) on the annotated set.

We applied the NER model to extract entities from within the predicted recommendation sentences. Table 8 shows the distribution of predicted entities by modality. As can be observed from the example sentences presented in Table 7, not all recommendation sentences included reason, test, or time frame information. From 871,680 recommendations, the NER model extracted 448,868 reason (51%), 777,618 test (89%), and 254,095 time frame (29%) entities.

| Imaging Modality | Number of recommendation sentences | Number of reports with recommendations (%) |
| --- | --- | --- |
| Angiography | 17,373 | 12,878 (24.00%) |
| Computed Tomography | 270,827 | 186,516 (26.38%) |
| Fluoroscopy | 182 | 172 (16.04%) |
| Magnetic Resonance Imaging | 80,484 | 54,482 (22.34%) |
| Mammogram | 213,846 | 155,335 (98.70%) |
| Nuclear Medicine | 17,149 | 11,967 (20.51%) |
| Portable Radiography | 19,934 | 18,734 (6.04%) |
| Positron emission tomography | 726 | 518 (28.79%) |
| Ultrasound | 135,049 | 108,126 (30.74%) |
| X-Ray | 116,110 | 109,575 (7.73%) |

**Table 6.** Number of predicted recommendation sentences by modality

| Imaging Modality | Example recommendation sentence |
| --- | --- |
| Angiography | *Patient will follow-up in four weeks for further evaluation.* |
| Computed Tomography | *However, follow-up CT is suggested to exclude possible recurrence of lymphoma* |
| Fluoroscopy | *Endoscopy or repeat CT scan may help to further evaluate this lesion.* |
| Magnetic Resonance Imaging | *Dedicated MRI of this region would be helpful for further evaluation.* |
| Mammogram | *Normal interval follow-up is recommended in 12 months.* |
| Nuclear Medicine | *CT of the chest without contrast could be used for further evaluation.* |
| Portable Radiography | *Recommend repeat CT scan of the chest for further evaluation of this lesion in 3 months.* |
| Positron emission tomography | *MRI is recommended for further evaluation.* |
| Ultrasound | *Four phase CT scan of the abdomen, with CT pelvis, is suggested for further evaluation of liver, abdomen and pelvis.* |
| X-Ray | *Shoulder views are recommended for further evaluation.* |

**Table 7.** Examples of recommendation sentences extracted from the dataset for each modality

| Imaging Modality | Reason | Test | Time frame |
| --- | --- | --- | --- |
| Angiography | 7,732 | 8,421 | 4,474 |
| Computed Tomography | 191,453 | 221,941 | 25,440 |
| Fluoroscopy | 159 | 125 | 7 |
| Magnetic Resonance Imaging | 41,136 | 68,452 | 20,679 |
| Mammogram | 24,998 | 250,605 | 162,421 |
| Nuclear Medicine | 11,895 | 12,476 | 974 |
| Portable Radiography | 15,292 | 15,725 | 367 |
| Positron emission tomography | 449 | 525 | 12 |
| Ultrasound | 82,371 | 134,233 | 36,827 |
| X-Ray | 73,383 | 65,115 | 2,894 |

**Table 8.** Number of predicted entities by modality

To understand the follow-up status of each identified recommendation, we performed a longitudinal analysis on the multi-institutional radiology dataset based on the information extracted by the NLP methods. To accomplish that, we first created a timeline of radiology reports for each patient based on the timestamps of reports in our dataset.

In our initial preliminary analysis, for each patient timeline we identified all reports with follow-up recommendations. We used the timestamps of the reports as the timestamps of the recommendations. For each identified recommendation, we checked whether a radiology test with the same modality occurred after the timestamp of the recommendation in the patient's timeline to roughly estimate the percentage of patients who stayed within the network of two hospitals in our dataset. Table 9 presents the results of this initial analysis.

| Imaging Modality | Reports with follow-up recommendation | No following tests of same modality | Had following tests of same modality |
|---|---|---|---|
| Angiography | 12,878 | 5534 (0.43%) | 7344 (0.57%) |
| Computed Tomography | 186,516 | 59659 (0.32%) | 126857 (0.68%) |
| Fluoroscopy | 172 | 144 (0.84%) | 28 (0.16%) |
| Magnetic Resonance Imaging | 54,482 | 27529 (0.51%) | 26953 (0.49%) |
| Mammogram | 155,335 | 45632 (0.29%) | 109703 (0.71%) |
| Nuclear Medicine | 11,967 | 7244 (0.61%) | 4723 (0.39%) |
| Portable Radiography | 18,734 | 5162 (0.28%) | 13572 (0.72%) |
| Positron emission tomography | 518 | 401 (0.77%) | 117 (0.23%) |
| Ultrasound | 108,126 | 41949 (0.39%) | 66177 (0.61%) |
| X-Ray | 109,575 | 32048 (0.29%) | 77527 (0.71%) |

**Table 9.** Number of patients who had follow-up / did not have follow-up

Next, we used the entities extracted by our NLP methods. We first identified all reports that had recommendation with a time frame entity. The text segments of the time frame entities were then normalized to a common temporal expression using the Stanford temporal tagger (SUTime)[31]. SUTime normalizes the temporal phrases into a value (e.g., *3 months* = P3M, *1 year* = P1Y). Then using the timestamp of the recommendation and the normalized time frame value for follow-up, we projected the next radiologic test date for the patient. Because some projected dates are outside of the collected time range of the dataset, we considered those radiology encounters censored (18,338 records). If the recommended time consists of a range such as "*6 to 12 months*", we used the end of the range to project the next visit. Furthermore, a report could contain multiple follow-up recommendations (35,375 records). If the patient did not have any one of the follow-up encounters as recommended in the report, we considered no follow-up for that report. If the patient was late to any one of the recommended follow-up encounters in the report, we considered late follow-up for that report. Table 10 shows the number of patients who did not have a follow-up encounter as recommended by radiologist as well as those who had a follow-up earlier or later than the recommended time.

| Imaging Modality | Reports with recommendation and normalized time frame | No follow-up | Early follow-up | Late follow-up |
|---|---|---|---|---|
| Angiography | 3,187 | 1198 (0.38%) | 570 (0.18%) | 1419 (0.45%) |
| Computed Tomography | 15,591 | 5849 (0.38%) | 4999 (0.32%) | 4743 (0.30%) |
| Fluoroscopy | 8 | 6 (0.75%) | 0 (0.00%) | 2 (0.25%) |
| Magnetic Resonance Imaging | 9,158 | 3641 (0.40%) | 1788 (0.20%) | 3729 (0.41%) |
| **Mammogram** | **122,223** | **27789 (0.23%)** | **20010 (0.16%)** | **74424 (0.61%)** |
| Nuclear Medicine | 399 | 178 (0.45%) | 132 (0.33%) | 89 (0.22%) |
| Portable Radiography | 375 | 201 (0.54%) | 80 (0.21%) | 94 (0.25%) |
| Positron emission | 9 | 7 (0.78%) | 1 (0.11%) | 1 (0.11%) |
| Ultrasound | 21,832 | 8855 (0.41%) | 5236 (0.24%) | 7741 (0.35%) |
| X-Ray | 1,820 | 761 (0.42%) | 333 (0.18%) | 726 (0.40%) |

**Table 10.** Number of patients who had no follow-up / early follow-up / late follow-up

As can be observed from Table 10, mammograms had the highest follow-up rate (77%: 16% early, 61% late follow-up). This is expected as mammograms are commonly used as a screening tool to detect early breast cancer in women and annual exam is recommended for women over 40yo. For the other modalities, the follow-up rates varied between 22% (positron emission tomography) and 62% (angiography, computed tomography).

## Conclusion

The main contribution of this paper is the application of deep learning to identify clinically important recommendation information in radiology notes. We applied the trained models to multi-institutional dataset of 3.3 million radiology notes and presented our very preliminary analysis of recommendation follow-up adherence over a period of 10 years.

One of the limitations of our study was the size of the training set for recommendations. To achieve good performance, deep learning approaches require relatively larger dataset than traditional machine learning methods. Our labeled training corpus consists of only 567 reports. The presented performance results are very promising. However, there is still room for improvement in recommendation extraction as well as NER tasks for reason, test, and time frame. We plan to annotate more reports to increase the size of our training set.

In our error analysis, we found that some of the time frame entities could not be normalized by Stanford's temporal tagger, such as "second trimester" in the recommendation "*Follow-up ultrasound is recommended in the early second trimester for further evaluation.*". 254,095 recommendations (29% of all recommendations) had time-frame entities and 174,602 (20% of all recommendations) of those with normalized time frames were included in the preliminary analysis. To utilize our entire dataset, we will build our own normalization algorithm for time frame entities. Additionally, we will automatically learn the recommended time frames for each modality from the data and use this knowledge to fill the missing time information for recommendations without time frame entities.

Our analysis did not utilize the extracted test and reason entities. We assumed the recommended test would be the same modality of the original test with recommendation mentioned in its report. However in reality, recommended test may be of a different modality or of the same modality but with a different anatomy. In future work, we will extract the recommended anatomy in addition to other entity types. In addition, test, anatomy, and reason entities will be mapped to standardized dictionaries to enable a more comprehensive follow-up adherence analysis.

## Acknowledgements

This publication was supported by the National Center For Advancing Translational Sciences of the National Institutes of Health under Award Number UL1 TR002319.